\documentclass[runningheads]{llncs}
\pdfoutput=1
\usepackage{booktabs} % For formal tables
\usepackage{mathtools}

\DeclarePairedDelimiter\floor{\lfloor}{\rfloor}
\usepackage{comment}
\usepackage{smartdiagram}
\usepackage{tikz}
\usesmartdiagramlibrary{additions}
\usepackage{xcolor}
\usepackage{csquotes} %for quotation marks
\newcommand{\WK}[1]{\textcolor{blue}{\small{(\textbf{WK:} \textit{#1})}}}
\newcommand{\SB}[1]{\textcolor{magenta}{\small{(\textbf{SB:} \textit{#1})}}}
\usepackage[draft]{changes}			% \added, \replaced, \deleted. With [final] instead of [draft] all changes are accepted
\definechangesauthor[name=WK, color=blue]{1}
\definechangesauthor[name=SB, color=orange]{2}
\setauthormarkuptext{name}			% superscript "WK" instead of superscript "1"
\usepackage{algorithm}
\usepackage{algorithmic}
\usepackage{amsmath} % some math-related packages, not sure which of them are necessary
\usepackage{amssymb}
\usepackage{amsfonts}
\newcommand{\mat}[1]{\mathbf{#1}}
\usepackage{graphicx}
\begin{document}
\title{SACOBRA with Online Whitening for Solving Optimization Problems with High Conditioning
\\\vspace{0.3cm} \small Technical Report}
\author{Samineh Bagheri\inst{1}\and
Wolfgang Konen\inst{1} \and
Thomas B\"ack\inst{2}}
\authorrunning{S. Bagheri et al.}
\institute{TH K\"oln -- Univeristy of Applied Sciences, Gummersbach, Germany \email{\{samineh.bagheri,wolfgang.konen\}@th-koeln.de}\and
Leiden University, LIACS, Leiden, The Netherlands
\email{t.h.w.baeck@liacs.leidenuniv.nl}\\
}
\maketitle              % typeset the header of the contribution
\begin{abstract}

Real-world optimization problems often have expensive objective functions in terms of cost and time. It is desirable to find near-optimal solutions with very few function evaluations. Surrogate-assisted optimizers tend to reduce the required number of function evaluations by replacing the real function with an efficient mathematical model built on few evaluated points. 
Problems with a high condition number are a challenge for many surrogate-assisted optimizers including SACOBRA.
To address such problems we propose a new online whitening operating in the black-box optimization paradigm.
We show on a set of high-conditioning functions that online whitening tackles SACOBRA's early stagnation issue and reduces the optimization error by a factor between $10$ to $10^{12}$ as compared to the plain SACOBRA, though it imposes many extra function evaluations.
Covariance matrix adaptation evolution strategy (CMA-ES) has for very high numbers of function evaluations even lower errors, whereas SACOBRA performs better in the expensive setting ($\leq 10^3$ function evaluations).
If we count all parallelizable function evaluations (population evaluation in CMA-ES, online whitening in our approach) as one iteration, then both algorithms have comparable strength even on the long run. This holds for problems with dimension $D \leq 20$.

\end{abstract}
%
%
%

%%%%%%%%%%%%%%%%%%%%%%%%%%%%%%%%%%%%%%%%%%%%%%%%
\section{Introduction}

Optimization problems can often be defined as minimization of a black-box objective function $f(\vec{x})$. An optimization problem is called black-box if no analytical information about itself or its derivatives are given. 
Evolutionary algorithms including covariance matrix adaptation evolution strategy (CMA-ES)~\cite{Hansen1996}, genetic algorithm (GA)~\cite{Sawyerr2013}, differential evolution (DE)~\cite{Posik2012}, and particle swarm optimization (PSO)~\cite{Saxena2015} are among strong derivative-free algorithms suitable for handling black-box optimization problems. All the mentioned optimization algorithms are inspired from the evolution theory of Darwin and tend to evolve a randomly generated initial population by means of different optimization operators (crossover, mutation, selection, estimating distribution etc.) iteratively. Despite all the significant contributions of differential evolution, solving problems with high-conditioning remains a challenge, as it is mentioned in~\cite{Sutton2007de}. In~\cite{Sawyerr2015} a genetic algorithm is evaluated on a set of black-box problems and it is observed that the algorithm is weak in optimizing high conditioning problems.  Despite many evolutionary-based algorithms, CMA-ES is very successful in tackling high-conditioning problems. The advantage of CMA-ES when solving problems with high conditioning stems from the fact that in each iteration the covariance matrix of the new distribution is adapted according to the evolution path which is the direction with highest expected progress. In other words, the covariance matrix adaptation aims to learn the Hessian matrix of the function in an iterative way.

Although the contribution of the mentioned evolutionary based algorithms is significant, they often require too many function evaluations which are not affordable in many real-world applications. That is because determining the value of the objective functions at a specific point $\vec{x}$ (set of variables) often requires to conduct a time-expensive simulation run. In order to solve expensive optimization problems in an efficient manner, several algorithms were developed which aim at reducing the number of function evaluations through the assistance of surrogate models~\cite{Bagheri2015a,Regis2015,Jones1998}. 

\begin{figure}%
\vspace*{0.6cm}
\resizebox{0.99\columnwidth}{!}{
\smartdiagramset{
module minimum width=2.5cm,
module x sep=3.2,
back arrow disabled,
text width =2.2cm,
set color list={magenta!25,teal!40,teal!40}
}
%\begin{center}
\smartdiagramadd[flow diagram:horizontal]{Initialization ,phase I: modeling, phase II: optimization}{}
\begin{tikzpicture}[overlay]
\draw[additional item arrow type,color=teal!60] (module3)-- ++(0,1) -| (module2);
\end{tikzpicture}
%\end{center}
}
\caption{Conceptualization flowchart of surrogate-assisted optimization}%
\vspace*{-0.6cm}
\label{fig:concept}%
\end{figure}
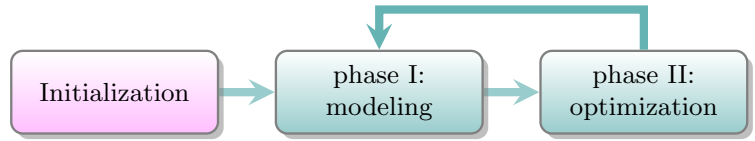

Many of the recently developed surrogate-assisted optimization algorithms go -- after an initialization step -- through two main phases shown in Fig.~\ref{fig:concept}. Phase I builds a cheap and fast mathematical model (surrogate) from the evaluated points. Phase II runs the optimization procedure \textit{on the surrogate} to suggest a new infill point. The algorithm is sequential: as soon as the new infill point is evaluated on the real function, it will be added to the population of evaluated points and the surrogate will be updated accordingly. The two phases are repeated until a predefined budget of function evaluations is exhausted.

 Clearly, the modeling phase has a significant impact on the performance of the optimizer. The surrogate-assisted optimization algorithm can be of no use, if the surrogate models are not accurate enough and do not lead the search to the interesting region.
 %The surrogate-assisted optimization algorithm can be of no use, if the surrogate models are not reasonably well-defined and do not lead the search to the interesting region.
 %\WK{What do you mean by well-defined?}\SB{Honestly I'm not sure anymore, so I reformulate this part a little bit}
 Therefore, it is very important to have an eye on the quality of the surrogates. Radial basis function interpolation (RBF) and Gaussian process (GP) models are commonly used for efficient optimization~\cite{Bagheri2017c,Jones1998,Bagheri2017b,Bagheri2017,Bhattacharjee2016,Loshchilov2012a}. 
 Although the mentioned techniques are suitable for modeling complicated non-linear functions, both may face challenges in handling other aspects of functions. 
SACOBRA~\cite{Bagheri2017} is an optimization framework which uses RBFs as modeling technique.
 This algorithm is very successful in handling the commonly used constrained optimization problems, the so-called G-function benchmark~\cite{CEC2006}.

 However, it performs poorly when optimizing functions with a large condition number. A function, that has a high ratio of steepest slope in one direction to flattest slope in another direction,  has a large condition number. We call this a function with \textit{high conditioning}.  The condition number of a function can be determined as the ratio of the largest to smallest singular value of its Hessian matrix.

Shir et al.\cite{Shir2011,Shir2014efficient} observe that in high-conditioning problems CMA-ES may converge to the global optimum but fail to learn the Hessian matrix. They propose with FOCAL an efficient approach for determining the Hessian matrix even for functions with high condition number.

The surrogate-assisted CMA-ES algorithms proposed in~\cite{Loshchilov2012a,Bajer2015} use surrogates in a different way: Whole CMA-ES generations alternate between being generated on the real function or on the surrogate function. Which function is used is determined by the algorithm online during the optimization run, based on a certain accuracy criterion. It turns out that for high-conditioning functions the algorithm effectively uses only the real function. Thus it behaves equivalent to plain CMA-ES and does not use surrogates in the high-conditioning case.

This work focuses on surrogate-assisted optimization of functions with moderate or high condition numbers. In Sec.~\ref{sec:problemstatement}, we provide  some illustrative insights \textit{why} such functions are tricky to optimize with surrogate-assisted solvers due to modeling difficulties. 
%As an illustration, we describe a function with high conditioning and the underlying issues for determining an accurate model for it in Section~\ref{sec:problemstatement}. 
Sec.~\ref{sec:methods} gives a brief description of the SACOBRA algorithm. Then we describe the newly proposed online whitening scheme added to SACOBRA for boosting up the model performance. The experimental setup and the results on 
%a subset of 
the noiseless single-objective BBOB benchmark~\cite{Fink2009} are described in Sec.~\ref{sec:exp} and \ref{sec:res}, resp. 
%Moreover, we show and discuss the performance of our algorithm on a subset of BBOB problems in Section~\ref{sec:res}. 
%The BBOB benchmark is a set of single-objective noiseless optimization problems described in~\cite{Fink09}. 
%This benchmark is very well-studied and often used to evaluate black-box optimizers. 
%Furthermore, we will discuss the results in Section~\ref{sec:res}. 
Sec.~\ref{sec:conc} concludes. % this work.

%%%%%%%%%%%%%%%%%%%%%%%%%%%%%%%%%%%%%%%%%%%%%%%%
%\section{Problem Statement}
\section{Why High Conditioning Is A Problem For Surrogates} % /WK/ a suggestion, what do you think? /SB/ sounds good
\label{sec:problemstatement}

In order to investigate the behavior of the RBF interpolation technique for modeling functions with high conditioning, we take a closer look at the second function $F02$ from the BBOB benchmark~\cite{Fink2009}: 
\begin{eqnarray}
 F_{02}(\vec{x})=    & \sum\limits_{i=1}^{D}\alpha_i z_i^2 =& \sum\limits_{i=1}^{D}10^{6\frac{i-1}{D-1}} z_i^2
            %\\& \mathbf{\alpha}^T \cdot \mathbf{z}^2             =& \begin{bmatrix}1 & \cdots & 10^6\end{bmatrix} \begin{bmatrix}z_1^2\\\vdots\\z_D^2\end{bmatrix},
\label{equ:f02}
\end{eqnarray}
where $\vec{z}=T_{osz}(\vec{x}-\vec{x}^*)$  and $T_{osz}(\vec{x})$ is a nonlinear transformation~\cite{Fink2009}, used to make the surface of $F_{02}(\vec{x})$ uneven without adding any extra local optima. 
%$F_{02}(\vec{x})$ is an ellipsoidal function with high conditioning.

%As it is formulated in Eq~\eqref{equ:f02}, $F_{02}(\vec{x})$ is the summation of the weighted squared variables (an elliposidal function). %\WK{Sentence not fully clear.} 
This function can be defined in any $D$-dimensional space. 
%The corresponding weight of the highest variable is always $10^6$ and the lowest variable is $1$. The variables between the lowest to highest have weights between 1 to $10^6$. 
The large difference between the weights of the lowest variable $x_1$ to the highest $x_D$ results in the high condition number of $10^6$.

\begin{figure}[!t]
\centering
\begin{tabular}{ll}
\hspace*{-0.4cm}
\begin{minipage}{0.5\columnwidth}
\includegraphics[width=1.0\textwidth]{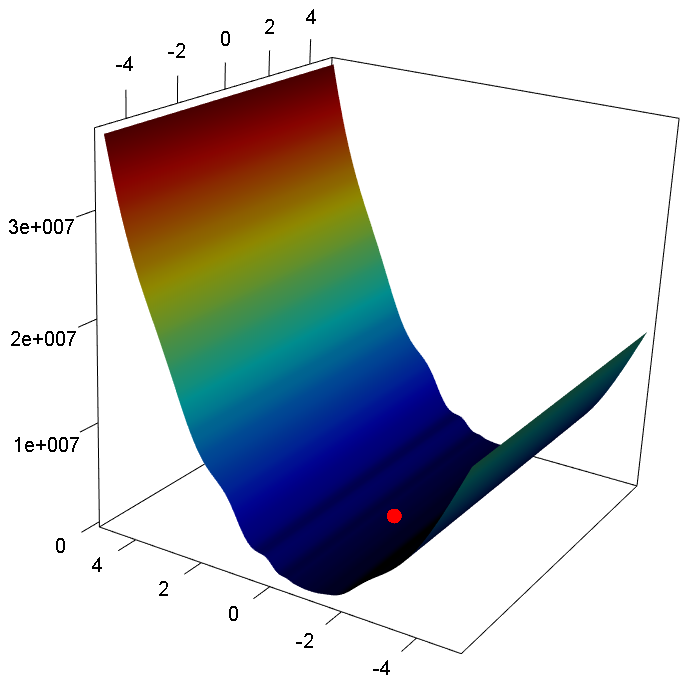}
\end{minipage}
& 
\hspace*{-0.45cm}
\begin{minipage}{0.5\columnwidth}
\includegraphics[width=1.0\textwidth]{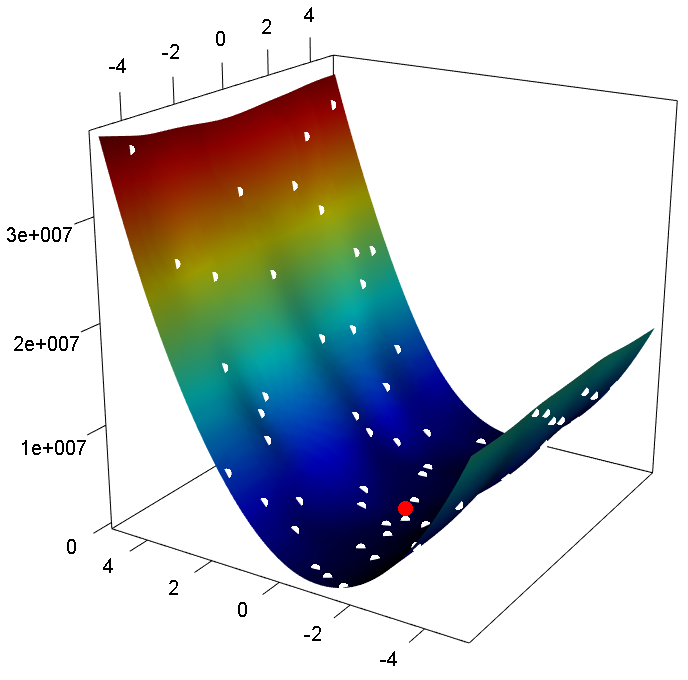}
%\WK{made the plots slightly bigger}
\end{minipage}
\end{tabular}
\caption{$F02$ function from the BBOB benchmark %set
(ellipsoidal function). Left: The real function. Right: RBF model for $F02$ built from 60 points (white points). The red point shows the location of the optimal solution.}
\label{fig:f02}
\end{figure}

Fig.~\ref{fig:f02}, left, shows how $F_{02}(\vec{x})$ looks like for D=2.
%a 2-dimensional function $F_{02}(\vec{x})$ looks like. 
It is easy to see that $F_{02}(\vec{x})$ has steep walls in one direction but looks pretty flat in the other direction. Fig.~\ref{fig:f02}, right, is the surrogate determined with a cubic RBF on 60 points (white dots).\footnote{We note in passing that a GP model for $F02$ would look structurally very similar.} We can see that the steep walls are reasonably well modeled but the surface is pretty wiggled. At first glance, it is not clear where the weakness of such model is.

\begin{figure}[!t]
\centering
\vspace{-0.3cm}
\includegraphics[width=0.99\columnwidth]{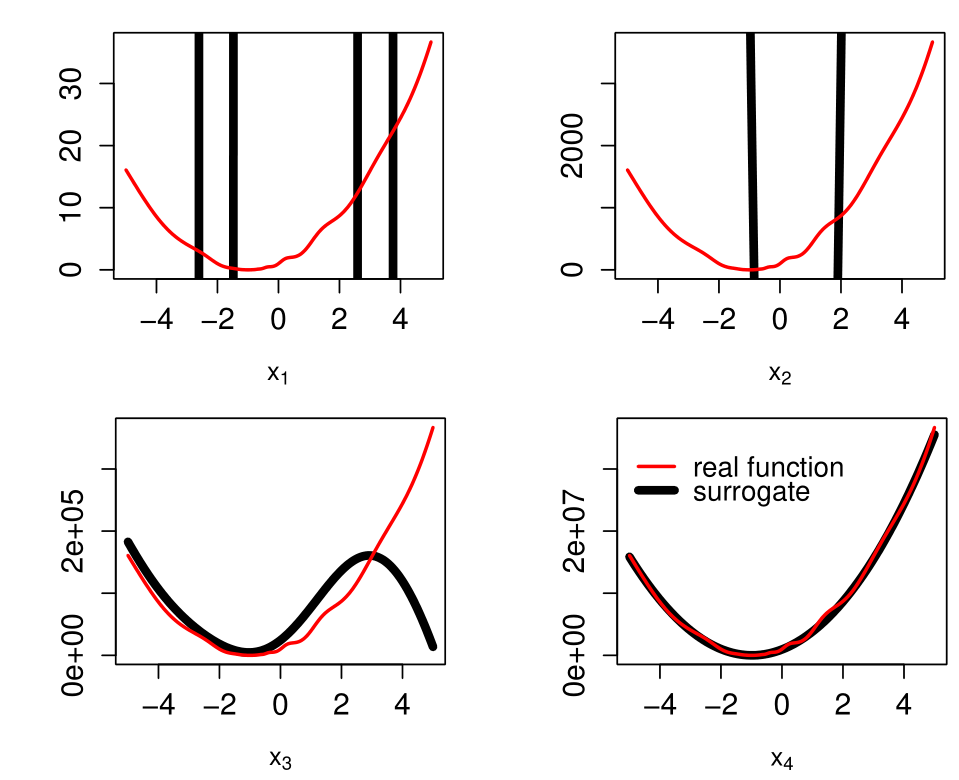}
\vspace{-0.45cm}
\caption{Four cuts at the optimum $\vec{x}^*$ of the 4-dimensional function $F02$ (Eq.~\eqref{equ:f02}) along each dimension. The red curve shows the real function and the black curve is the surrogate model. The black curve follows the red curve only in the 'steep' dimension $x_4$ (and to some extent in dimension $x_3$). Note the varying y-scales.
}
\label{fig:ps}
\end{figure}

In order to have a closer insight and also to be able to visualize higher-dimensional versions of $F_{02}(\vec{x})$ we plot cuts of the function along each dimension. Fig.~\ref{fig:ps} shows four cuts of $F_{02}(\vec{x})$ in the case $D=4$ where $\vec{x}$ is a 4-dimensional vector. In this example the optimum is at $x^*=(-1,-1,-1,-1)$.

As one can see, the highest dimension $x_4$ with the largest coefficient $\alpha_4=10^6$ is very well modeled, but the model slices for lower dimensions do not follow the real function and do not contain any useful information about the location of the optimum.

It is important to mention that what makes $F_{02}(\vec{x})$ a function challenging to model is not the large or small coefficients for each dimension but the large variations of steepness in different directions. 

Optimizing the surrogate model shown in Fig.~\ref{fig:ps} will result in a point $\vec{x}_{new}$, which has a near-optimal value for the steepest dimension but pretty much random values in all other dimensions.

%%%%%%%%%%%%%%%%%%%%%%%%%%%%%%%%%%%%%%%%%%%%%%%%
\section{Methods}
\label{sec:methods}
This work was 
%originally 
motivated by applying the SACOBRA optimizer to the single-objective BBOB set of problems. 
%and dealing with that fact that SACOBRA performs poor on many of those problems. 
Although we initially learned that SACOBRA performs poorly on problems with high and moderate conditioning, we investigated the underlying reason and came up with a cure: the so-called online whitening scheme. 
%This Section first describes SACOBRA briefly and then the proposed solution.

\subsection{SACOBRA: \textbf{S}elf-\textbf{A}djusting \textbf{C}onstrained %Black-Box 
\textbf{O}ptimization \textbf{B}y \textbf{R}BF \textbf{A}pproximation}   
SACOBRA, an extension of the COBRA algorithm~\cite{Regis2014}, is a surrogate-assisted optimizer originally designed for high-dimensional constrained black box optimization problems~\cite{Bagheri2017}, but also applicable to unconstrained problems.
SACOBRA uses augmented RBF models (Sec.~\ref{sec:augmentRBF}) as surrogates and then applies a constraint optimizer to solve the optimization problem on the model(s) (phase II in Fig.~\ref{fig:concept}). The optimization result is evaluated on the real function and is added to the population of points. Then, the model(s) will be updated (phase I) and the former steps will be repeated as long as the budget is not exhausted. SACOBRA uses self-adjusting techniques to tune sensitive parameters automatically~\cite{Bagheri2017}. Although SACOBRA appears to be strong in solving G-problems~\cite{CEC2006}, it is weak on optimizing functions with high conditioning, mainly due to the modeling phase.

\subsection{Augmented RBF}
\label{sec:augmentRBF}

Augmented RBFs are  linearly weighted combinations of radial basis functions and a polynomial tail as follows:

\begin{equation}
 \hat{f}(\vec{x})= \sum_{i=1}^n \theta_i \varphi(||\vec{x}       - \vec{x}_{(i)}||) + p(\vec{x}), \quad\vec{x} \in \mathbb{R}^D,
\label{eq:RBFfunction}
\end{equation}

where $\{\vec{x}_{(i)} \in \mathbb{R}^D| i=1,...n\}$ is the current SACOBRA population and $p(\vec{x}) = \mu_0 + \vec{\mu}_1\,\vec{x}+\vec{\mu}_2\,\vec{x}^2 \cdots+\vec{\mu}_k\,\vec{x}^k$ is a $k$-th order polynomial in $D$ variables with $kD+1$ coefficients. 

The augmented RBF model requires the solution of the following linear system of equations:
\begin{equation}
\begin{bmatrix} \mat{\Phi} & \mat{P} \\ \mat{P}^T & \mat{0} \end{bmatrix}  %\mat{0}_{(kd+1)\times(kd+1)}
\begin{bmatrix} \vec{\theta} \\ \vec{\mu}\,' \end{bmatrix}  = \left[ \begin{array}{c} \vec{f} \\ \vec{0} \end{array} \right]
\label{eq:RBF}
\end{equation}

Here, $\mat{\Phi} \in \mathbb{R}^{n\times n}$ with $\mat{\Phi}_{ij} = \varphi(||\vec{x}_{(j)}- \vec{x}_{(i)}||)$ and $\mat{P} \in \mathbb{R}^{n\times(kD+1)}$ is a matrix with $(1,\vec{x}_{(i)},\ldots,\vec{x}_{(i)}^k)$ in its $i$th row. $\mat{0} \in \mathbb{R}^{(kD+1)\times(kD+1)}$ is a zero matrix, $\vec{f}$ is a vector with $f(\vec{x}_{(i)})$ in its $i$th component and $\vec{\mu}\,'$ is the concatenation of the polynomial coefficients in $p(x)$.

In this work we use $\varphi(r)=r^3$ (cubic radial basis functions) with a second order polynomial tail ($k=2$).

\subsection{Online Whitening}

As described in Section~\ref{sec:problemstatement}, functions with high conditioning are difficult to model for RBF or GP surrogates. 
%\WK{Do we have evidence (or can we provide evidence) that problems we have with RBF surrogates are also present for GP surrogates? And that Online Whitening would be of help in both cases? This would strengthen our argument.}\SB{Yes, we have. Fig~\ref{fig:EGO3} shows that after 500 iterations EGO converges to solutions with large error in range of $1e+3$} 
Although the overall modeling error may be small, the models often have spurious local minima along the 'shallow' directions. This obviously hinders optimization. What we show here for RBF surrogate models holds the same way for GP (or Kriging) surrogate models often used in EGO~\cite{Jones1998}: Problems with a high condition number have a much higher optimization error than those with low conditioning (differing by a factor of $10^7$ after 500 function evaluations, as some preliminary experiments have shown that we undertook with EGO using a Matern(3/2)-kernel).
%They often cannot provide proper models especially for optimization tasks, since they introduce spurious local minima. 

In order to tackle high-conditioning problems with surrogate-assisted optimizers, we propose the online whitening scheme described in Algorithm~\ref{alg:alg1}: We seek to transform the objective function $f(\vec{x})$ with high conditioning to another function $g(\vec{x})$ which is easier to model by surrogates:

\begin{equation}\label{eq:transformation0}
 g(\vec{x}) =f(\mathbf{M} (\vec{x}-\vec{x}_c)),
\end{equation}
\noindent
where $\mathbf{M}$ is a linear transformation matrix and $\vec{x}_c$ is the transformation center. The ideal transformation center is the optimum point which is clearly not available. As a substitute, we use in selected iterations the best so-far solution $\vec{x}_{best}$ as the transformation center. The transformation matrix $\mathbf{M}$ is chosen in such a way that the Hessian matrix of the new function becomes the identity matrix:

\begin{eqnarray}\label{eq:transformation1}
\frac{\partial^2 g(\vec{x})}{\partial \vec{x}^2} &=& \mathbf{I}
\end{eqnarray}
 
%\SB{see Section~\ref{sec:append} for derivation of the transformation matrix $\mathbf{M}$ }
It is derived in Appendix~\ref{sec:append} that  a solution for Eqs.~\eqref{eq:transformation0} and \eqref{eq:transformation1} is given by:

\begin{eqnarray}\label{eq:transformation2}
\mathbf{M}&=&\mathbf{H}^{-0.5}
\end{eqnarray}
where $\mathbf{H}$ denotes the Hessian matrix of the objective function $f$.

Appendix~\ref{sec:invsquareroot} shows how to calculate $\mathbf{M}$ in a numerically stable way. 
The transformation matrix $\mathbf{M}$ used in our proposed algorithm is similar to the so-called Mahalanobis whitening or sphering transformation, which is commonly used in statistical analysis~\cite{Kessy2017}. %\footnote{
A \textit{whitening or sphering transformation} aims at transforming a function in such a way that it has the same steepness in every direction, e.~g. the height map of an ellipsoidal function will become spherical.
%} 

After determining the transformation matrix, we evaluate all points in the population $\mathbf{X}$ on the new function $g(\vec{x})$ and store the pairs $\left(\vec{x}_{(k)}, g(\vec{x}_{(k)})\right)$ in the set $\mathbf{G}$ (steps 4 and 5 in Algorithm~\ref{alg:alg1}). Then we re-build the surrogate model for %the transformed function 
$g(\vec{x})$ by passing the set 
%the input-output pairs of
$\mathbf{G}$ to the RBF model builder (step 6).

%\WK{We need to describe how the Hessian is calculated or give at least a reference.}\SB{In the experimental setup section, I mention the numDeriv package}
%It is important to mention that, determining the Hessian matrix of the objective function \added{numerically} at each point, requires $4D+4D^2$ function evaluations which makes the proposed scheme costly. 
The Hessian matrix is determined numerically by means of Richardson's extrapolation~\cite{Bates1988} which requires $4D+4D^2$ function evaluations. 
%This makes the proposed scheme costly.
%However, our initial tests have shown that a frequent update of the Hessian matrix in each iteration of SACOBRA is not 
Initial tests have shown that an update of the Hessian matrix in each iteration of SACOBRA is not necessary. Thus we reduce the number of function evaluations by calling the online whitening scheme only every 10 iterations.

%The proposed algorithm
\begin{algorithm}[tbp]
\caption{Online whitening algorithm. Input: Function $f$ to minimize, population $\mathbf{X}=\left\{\vec{x}_{(k)}| k=1,\ldots, n\right\}$ of evaluated points, $\vec{x}_{best}$: best-so-far point from SACOBRA.
%\WK{Can there be numerical problems when calculating $\mathbf{H}^{-0.5}$? - Now clarified with Appendix B.}
%\WK{We should describe how we define and update $\vec{x}_{best}$}
}
\label{alg:alg1}
\begin{algorithmic}[1]
%\renewcommand{\algorithmicrequire}{\textbf{Input:}}
%\renewcommand{\algorithmicensure}{\textbf{Output:}}
 %\REQUIRE $f(\vec{x})$, $\mathbf{X}=\left\{\vec{x}_{(k)}| k=1,\ldots, n\right\}$, $\vec{x}_{best}$ 
 %\ENSURE  surrogate model $s(\vec{x})$ \\ 
%  \textit{Calculation} :
  \STATE $\mathbf{H} \leftarrow$ Hessian matrix of function $f(\vec{x})$ at $\vec{x}_{best}$
	\STATE $\mathbf{M} \leftarrow \mathbf{H}^{-0.5}$    \COMMENT{see Eq.~\eqref{eq:transformation2} and Appendix~\ref{sec:invsquareroot}}
	\STATE Update $\vec{x}_{best}$ with the function evaluations from Hessian calculation
	%\STATE Update $\vec{x}_{best}$ if any better solution is evaluated in Step.1
	\\ \textit{Transformation} :
  \STATE $g(\vec{x}) \leftarrow f(\mathbf{M} (\vec{x}-\vec{x}_{best}))$
	%\STATE $F_k \leftarrow g(\vec{x}_k)$ 
	\STATE $\mathbf{G} \leftarrow \left\{\left(\vec{x}_{(k)}, g(\vec{x}_{(k)})\right)|k=1, \ldots, n\right\}$  \COMMENT{evaluate all the points in $\mathbf{X}$ on the new function $g(\vec{x})$}
	%%%%% WK: I think that \mathfrak{G} is not well readable %%%%
	\STATE $s\left(\vec{x}\right) \leftarrow $ build surrogate model from $\mathbf{G}$
 \RETURN $s\left(\vec{x}\right)$   \COMMENT{surrogate model for next SACOBRA step}
\end{algorithmic}
\end{algorithm}

\begin{figure}[!t]
\centering
\includegraphics[width=0.99\columnwidth]{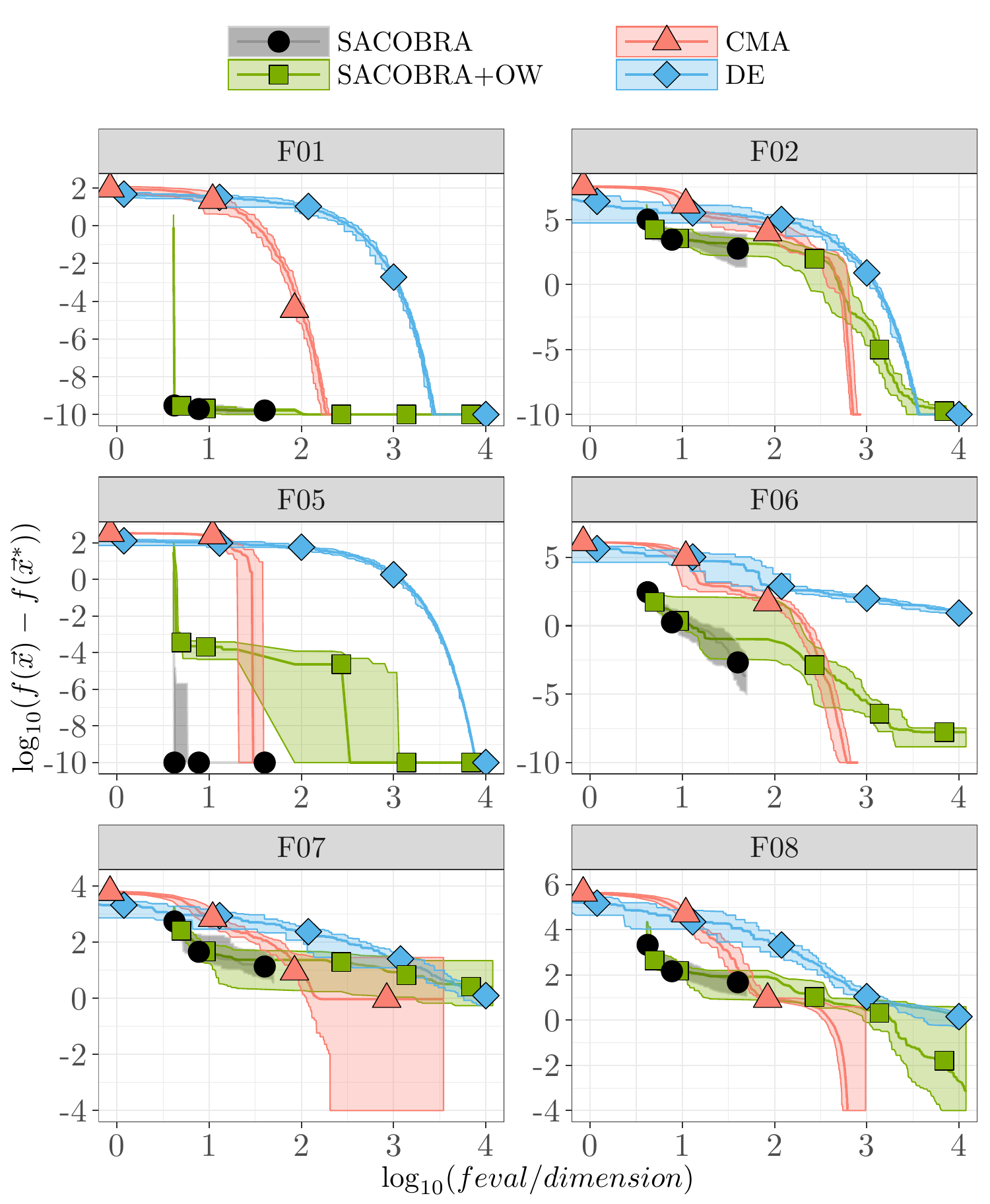}
\caption{Comparing the performance of SACOBRA, SACOBRA+OW, DE and CMA-ES  algorithms on F01, F02, F05, F06, F07 and F08 optimization problems ($D=10$).
}
\label{fig:res1}
\end{figure}

\begin{figure}[!t]
\centering
\includegraphics[width=0.99\columnwidth]{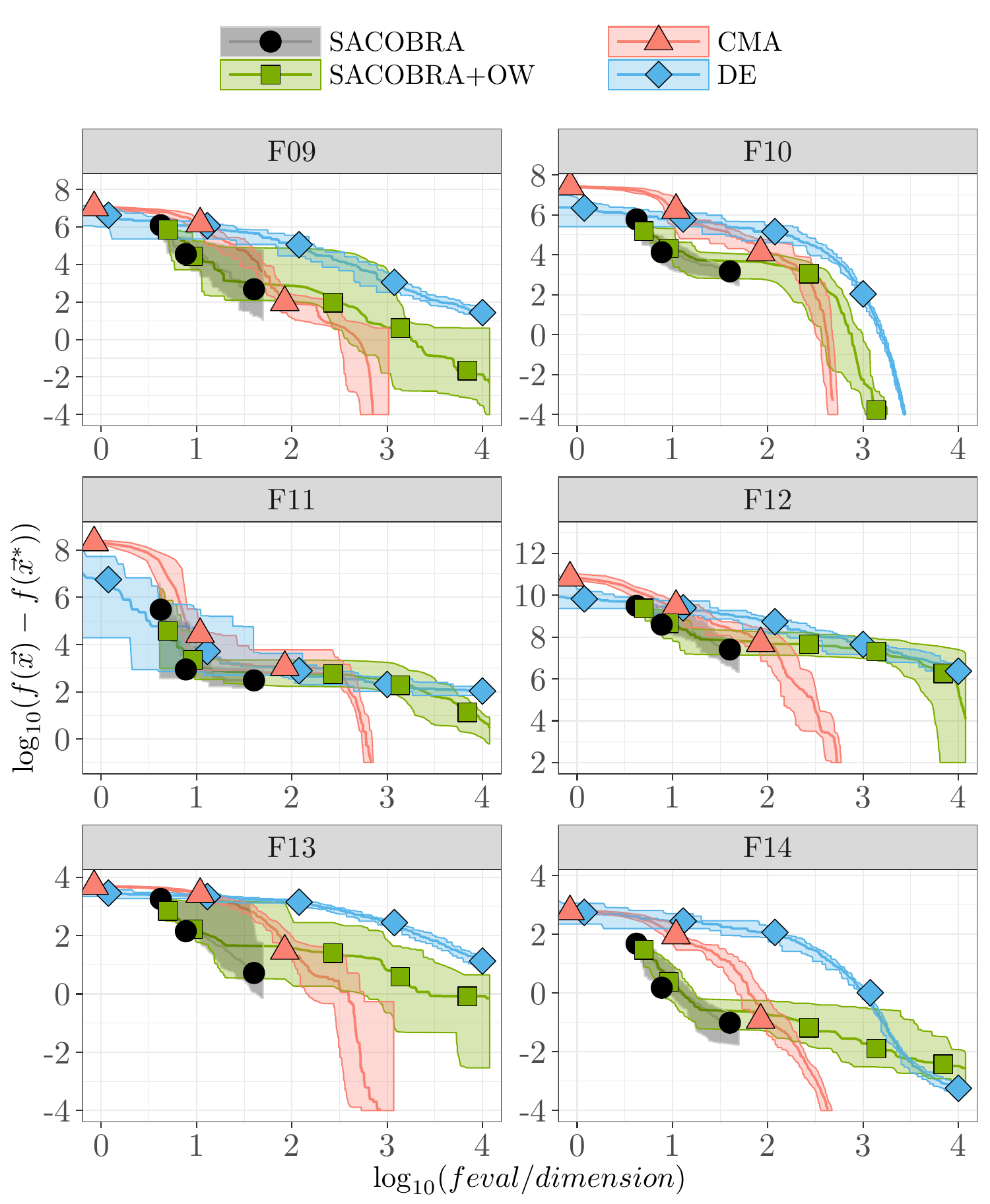}
\caption{Comparing the performance of SACOBRA, SACOBRA+OW, DE and CMA-ES  algorithms on F09, F10, F11, F12, F13 and F14 optimization problems ($D=10$).
}
\label{fig:res2}
\end{figure}

%%%%%%%%%%%%%%%%%%%%%%%%%%%%%%%%%%%%%%%%%%%%%%%%
\section{Experimental Setup}
\label{sec:exp}
We investigate the effectiveness of the online whitening scheme by comparing the standard SACOBRA algorithm (package \textsc{SACOBRA} in \texttt{R}) to SACOBRA combined with the online whitening scheme (SACOBRA+OW). To do so, we apply them to 12 of the 14 problems from the three first BBOB benchmarks~\cite{Fink2009}. We exclude two highly multimodal problems (F03 and F04), since they cannot be solved by surrogate modeling. Most of these benchmark functions have moderate to high condition numbers (see Table~\ref{tab:condNumbers}).

\begin{table}[b]%
\caption{Condition numbers for all the investigated problems. The condition number is defined as the ratio of slope in the steepest direction to the slope in the flattest direction~\cite{Hansen2011}.
}
\label{tab:condNumbers}
\centering
\begin{tabular}{l|c||l|c}
\toprule
\textbf{Function} & Condition number & \textbf{Function} & Condition number\\\hline\midrule
F01 &$1$ & 		F09 &$10^2$\\
F02 &$10^6$&	F10 &$10^6$\\
F05 &$1$ &		F11 &$10^6$\\
F06 &$10^3$&	F12 &$10^6$\\
F07 &$10^2$&	F13 &$10^2$\\
F08 &$10^2$&	F14 &$10^4$\\\bottomrule
\end{tabular}
\end{table}

Both algorithms, SACOBRA and SACOBRA+OW, are compared as well to the differential evolution (DE) algorithm~\cite{Price2005} and to the covariance matrix adaptation evolutionary strategy (CMA-ES)~\cite{Hansen1996}, using the \textsc{DEoptim} and \textsc{rCMA} packages in \texttt{R}, resp. Both optimizers are used with their standard parameters. The default population size is in this case  $10D$ and $4 + 3 \floor{\ln(D)}$ for the packages \textsc{DEoptim} and \textsc{rCMA}, respectively.   
%\WK{Should we mention here the most important parameters of DE and CMA-ES, i.e. population sizes and similar? (might be relevant for understanding the factor between fe and iteration)} \SB{I added the information}

The two surrogate-assisted algorithms (SACOBRA and SACOBRA+OW) have an initial population size of $4D$ individuals. A maximum population size of $50D$ is permitted for both SACOBRA algorithms. It is important to mention that SACOBRA+OW may evaluate more than one point per iteration. 

The online whitening scheme in SACOBRA+OW is first called after $20D$ iterations and it will be updated after each $10$ iterations. The numerical calculation of the Hessian matrix is performed with the numDeriv package in \texttt{R}. In this work we mainly study and present results for the 10-dimensional problems. In the end, we compare the  performance of all algorithms for 5- and 20-dimensional problems as well.

In order to compare the overall performance of different optimization algorithms on a set of problems we use data profiles~\cite{WildMore2009}:  
\begin{equation}
		d_{s}(\alpha)=\frac{1}{| \mathbb{P}|}|\{p \in  \mathbb{P} : \frac{t_{p,s}}{D_{p}} \leq \alpha\}|,
\label{eq:dprof}
\end{equation}
where $\mathbb{P}$ is a set of problems, $\mathbb{S}$ is a set of solvers and $t_{p,s}$ is the number of iterations that solver $s \in \mathbb{S}$ needs to \textit{solve} problem $p \in \mathbb{P}$. $D_p$ is the dimension of problem $p$. 
%$\alpha$ is a fixed performance ratio. \WK{I've added the def of $\alpha$ added at the end of this paragraph}
An optimization problem is said to be \textit{solved} if a solution $\vec{x}_{best}$ is found whose 
%optimization error is smaller than a given tolerance 
objective value $f(\vec{x}_{best})$ deviates from the true solution $f(\vec{x^*})$ less than a given tolerance $\tau$: 
\begin{equation}
		|f(\vec{x}_{best}) - f(\vec{x}^*)|<\tau
\end{equation}
Data profiles plot $d_{s}(\alpha)$ against $\alpha$ with $\alpha = \mbox{feval/dimension}$.

\begin{figure}[!t]
\centering
\includegraphics[width=0.99\columnwidth]{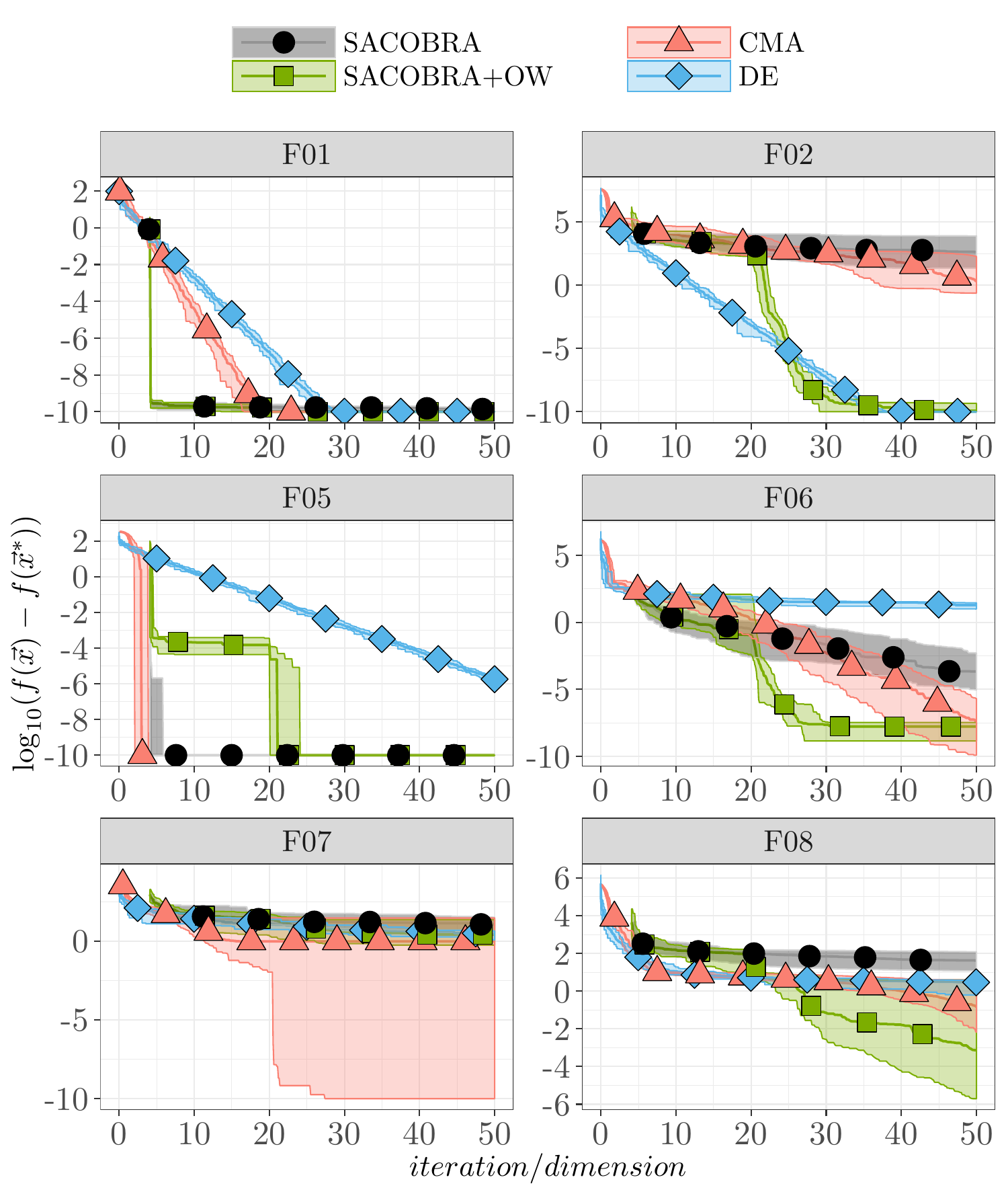}
\caption{Comparing the performance of SACOBRA, SACOBRA+OW, DE and CMA-ES  algorithms on F01, F02, F05, F06, F07 and F08 optimization problems ($D=10$). Now the x-axis shows iterations instead of function evaluations.
}
\label{fig:iter-res1}
\end{figure}

\begin{figure}[!t]
\centering
\includegraphics[width=0.99\columnwidth]{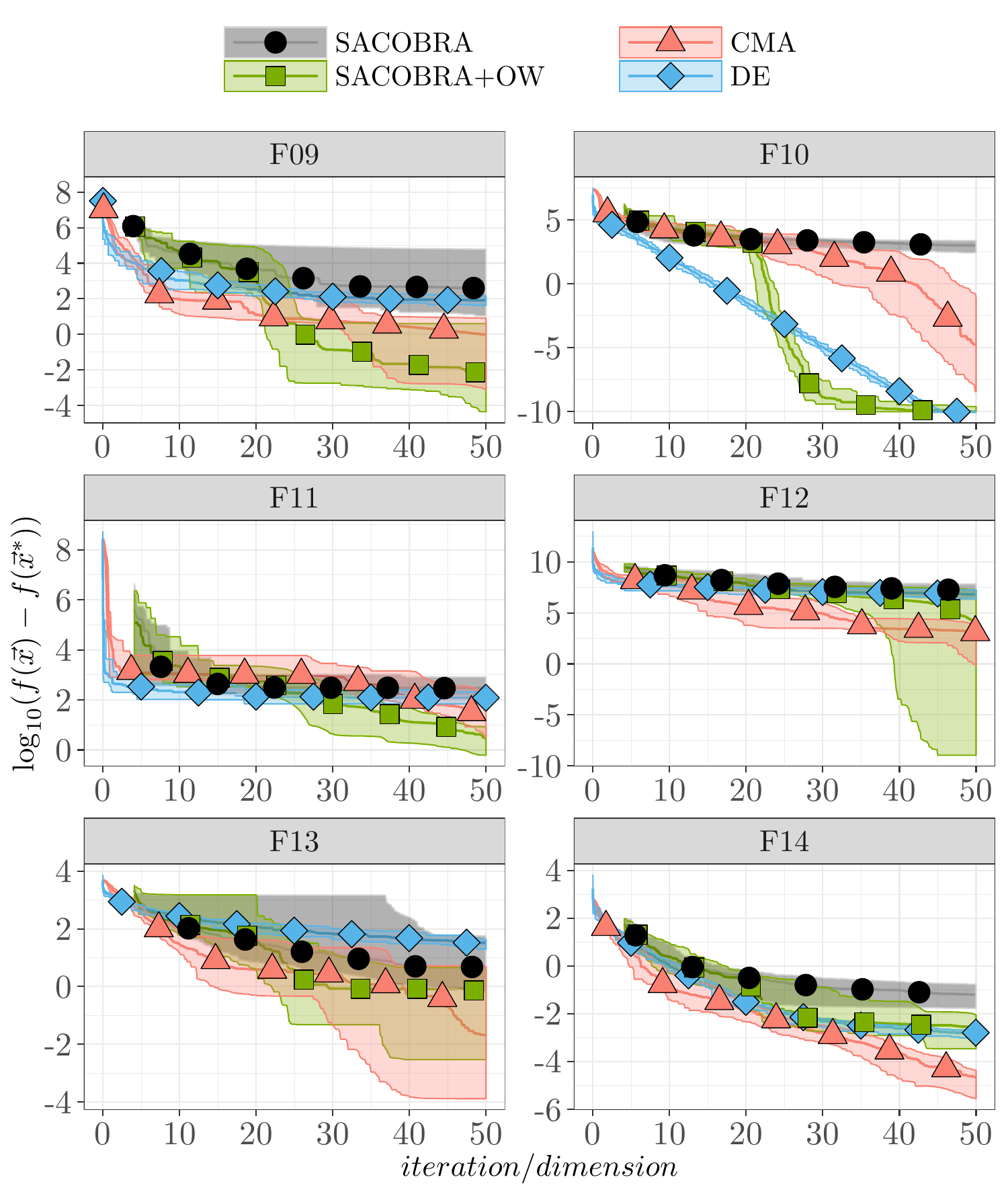}
\caption{Comparing the performance of SACOBRA, SACOBRA+OW, DE and CMA-ES  algorithms on F09, F10, F11, F12, F13 and F14 optimization problems ($D=10$). Now the x-axis shows iterations instead of function evaluations.
}
\label{fig:iter-res2}
\end{figure}

%%%%%%%%%%%%%%%%%%%%%%%%%%%%%%%%%%%%%%%%%%%%%%%%
\section{Results \& Discussion}\label{sec:res}

Figs.~\ref{fig:res1}--\ref{fig:res2} compare the optimization results achieved by SACOBRA, SACOBRA+OW, CMA-ES and DE on the BBOB benchmark problems listed in Table~\ref{tab:condNumbers}. 
%(excluding the multimodal problems F03 and F04). 
Both SACOBRA and SACOBRA+OW become computationally expensive as the population size grows. Therefore we apply them for at most $50D$ iterations on each problem. This is the reason why all SACOBRA curves in Figs.~\ref{fig:res1}--\ref{fig:res2} end at $1.7 = log_{10}(500/10)$ corresponding to a population size of 500. But SACOBRA+OW utilizes more real function evaluations when it starts to do the online whitening as described in Algorithm~\ref{alg:alg1}. 

SACOBRA solves problems with low conditioning like F01 (sphere function) and F05 (linear slope) after very few function evaluations ($<10D$) with a very high accuracy. CMA-ES and DE require 10 to 1000 times more function evaluations to find solutions as accurate as SACOBRA for these two problems. This strong performance of SACOBRA for F01 and F05 is probably due to the near-perfect models that can be built with RBFs for such simple functions from just a few points. 

However, for more complicated functions with high conditioning, SACOBRA often stagnates at a mediocre solution. 
Observing SACOBRA's behavior on high-conditioning functions in Figs.~\ref{fig:res1}--\ref{fig:res2} indicates that, although SACOBRA has a fast progress in the first 100 iterations, it gradually becomes very slow and eventually stagnates. This is because the surrogates model only the steep walls reasonably well. Therefore, after being down in the valley between the steep walls, SACOBRA is effectively blind for the correct direction, and it suggests random points within the valley. This picture makes it clear -- and experimental results confirm this -- that it is of no use to add more points to the SACOBRA population, because the  surrogate model stays wrong in all directions but the steepest ones.

SACOBRA+OW, which uses online whitening as a remedy for the modeling issues, can boost SACOBRA's optimization performance significantly. As it is shown in Figs.~\ref{fig:res1}--\ref{fig:res2}, SACOBRA+OW finds solutions whose optimization errors are  between $10$ times (in the case of F07) and $10^{12}$ times (in the case of F02) smaller than in SACOBRA. 

Although SACOBRA and SACOBRA+OW have the same population sizes, the latter requires significantly more function evaluations due to the Hessian calculation in the whitening procedure.

This makes SACOBRA+OW no longer suitable for expensive optimization benchmarks, if the real world restrictions does not permit any form of parallelization of the Hessian matrix computation.
%{This makes SACOBRA+OW no longer suitable for expensive optimization benchmarks.}

But it shows how to utilize surrogate models in cases with medium to high function evaluation budgets, which usually cannot be consumed completely by the surrogate model population.

Although SACOBRA+OW outperforms DE in 10 of 12 problems, it can compete with CMA-ES only when the  function evaluation budget is $10^3$ or less. Beyond this point, CMA-ES is usually the best algorithm.

Now we turn to the 'optimistic parallelizable' case: The numerical calculation of a Hessian matrix 
%of a function 
is not a sequential procedure and can be performed in parallel. Therefore, if enough computational resources are available, the Hessian matrix can be determined in the same time that a SACOBRA iteration needs. We call this the 'optimistic parallelizable' case. In this case, the efficiency of the SACOBRA+OW optimizer should be measured by its improvement per iteration (which need to be done one at a time). In the evolutionary strategies DE and CMA-ES, the evaluation of populations in each generation can be parallelized as well. So we count similarly all function evaluations needed to evaluate one DE- or CMA-ES-generation as \textit{one} iteration, in order to establish a fair comparison. 

Figs.~\ref{fig:iter-res1}--\ref{fig:iter-res2} depict the optimization error \textit{per iteration}\footnote{Each OW call is counted as one iteration, as well as each SACOBRA call. OW is first called at iteration $20D$
and then after each 10 SACOBRA iterations, one OW call is performed.} of SACOBRA, SACOBRA+OW, DE and CMA-ES %algorithms 
for the BBOB problems listed in Tab.~\ref{tab:condNumbers}. We compare the performances of the mentioned algorithms within the first 500 iterations. As illustrated in Fig.~\ref{fig:iter-res1}--\ref{fig:iter-res2}, SACOBRA+OW appears to be the leading algorithm in terms of speed of convergence for 8 of the problems. F07 and F14 are the only problems for which CMA-ES can find significantly better solutions than SACOBRA+OW within the limit of 500 iterations.  F05 and F13 can be optimized by CMA-ES and SACOBRA+OW similarly well. In general, SACOBRA+OW outperforms DE, although DE finds better solutions for F02 and F10 in the early iterations $1, \ldots, 250$ before SACOBRA+OW overtakes.

%\WK{A critical reviewer might ask: How exactly did you count the iterations in Fig.~\ref{fig:iter-res1}--\ref{fig:iter-res2}? -- First $20D$ SACOBRA-iterations, then 1 Hessian iteration, then 10 SACOBRA-iterations, then 1 Hessian, then 10 SACOBRA-iterations, and so on? (I think this is the way it should be done.)}\SB{Yes, so strictly speaking each OW is counted as one iteration}

\begin{figure}[!t]
\centering
\includegraphics[width=0.99\columnwidth]{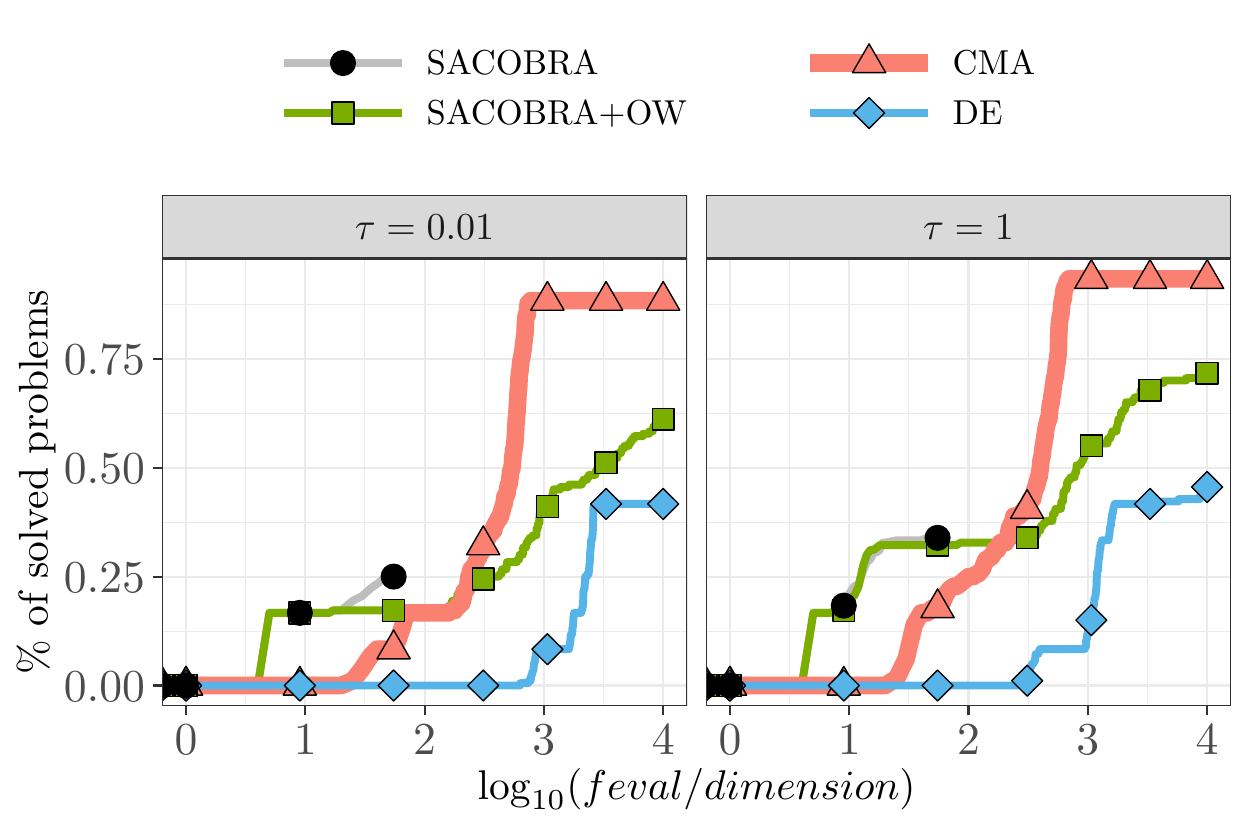}
\caption{Data profiles, Eq.~\eqref{eq:dprof}, for the algorithms SACOBRA, SACOBRA+OW, DE and CMA-ES, showing the overall performance on 12 BBOB problems with dimension $D=10$. The x-axis has the number of function evaluations, divided by $D$.
%\WK{A suggestion for a new figure: Does it make sense to show data profiles as a function of iterations/dimension?}\SB{I think it makes sense in case we assume parallel computing is possible, see Fig.~\ref{fig:DP-iter}}
}
\label{fig:DP}
\end{figure}

%Figs.~\ref{fig:DPexpensive} and 
Fig.~\ref{fig:DP} compares the overall performance of the four investigated algorithms by means of data profiles (Sec.~\ref{sec:exp}). %Fig.~\ref{fig:DPexpensive} 
It shows that the surrogate-assisted optimization is superior for low budgets (up to $100D$ function evaluations).
%, and Fig.~\ref{fig:DP} reveals that this advantage continues up to $100D$.

Fig.~\ref{fig:DP} indicates that SACOBRA can only solve $25\%$ of the problems with accuracy $\tau=0.01$, while SACOBRA+OW increases this ratio to about $62\%$. With the same accuracy level, our proposed algorithm can solve $25\%$ more problems than DE but also about  $25\%$ less than CMA-ES.

\begin{figure}[!t]
\centering
\includegraphics[width=0.99\columnwidth]{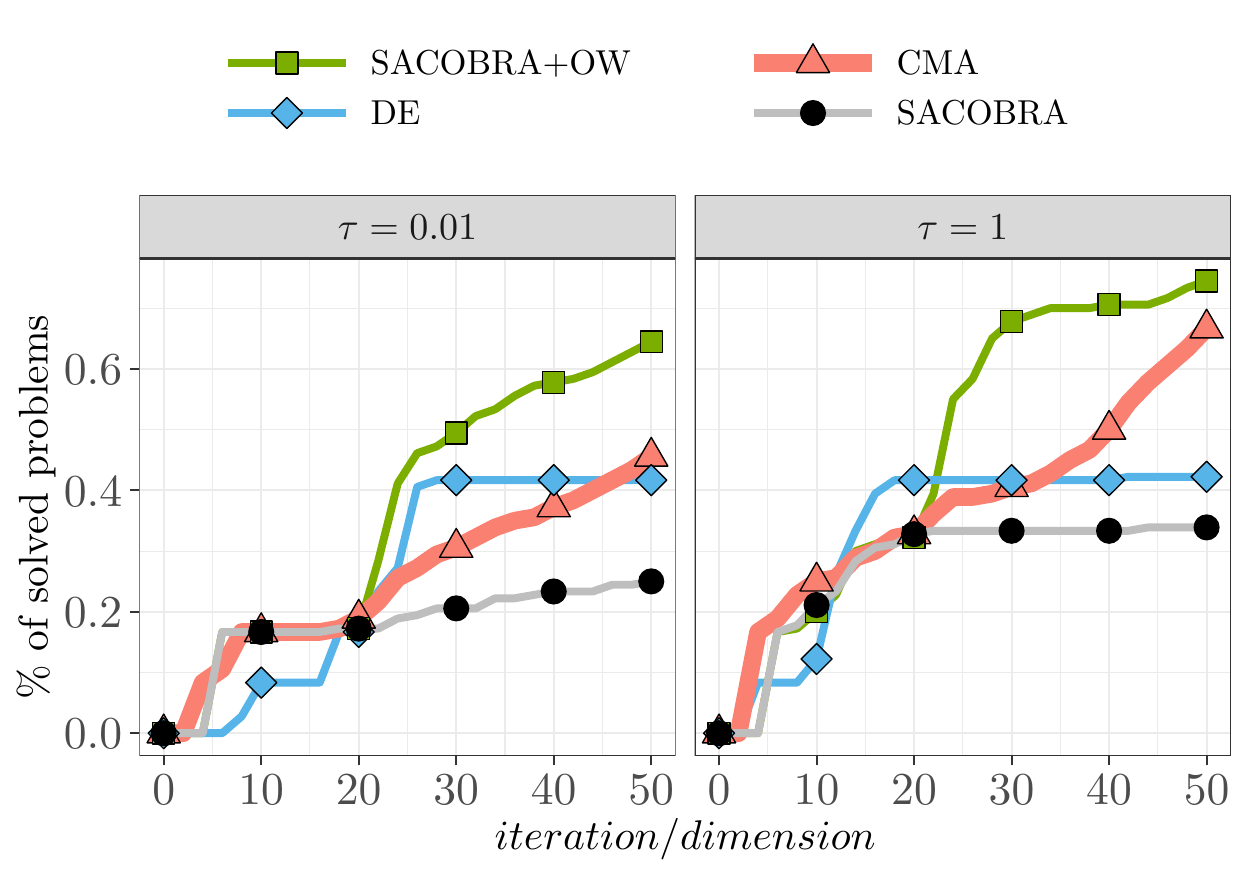}
\caption{
Same as Fig.~\ref{fig:DP}, but now for the 'optimistic parallelizable' case: We show on the x-axis the number of iterations (or generations), divided by $D$.
%Comparing the overall performance of SACOBRA, SACOBRA+OW, DE and CMAES algorithms on the 12 studied problems with $D=10$.
}
\label{fig:DP-iter}
\end{figure}

Fig.~\ref{fig:DP-iter} shows the data profiles for the 'optimistic parallelizable' case. Here SACOBRA+OW is consistently better than all other algorithms if we spent a budget of at most $50D$ iterations.  

\begin{figure}[!t]
\centering
\includegraphics[width=0.99\columnwidth]{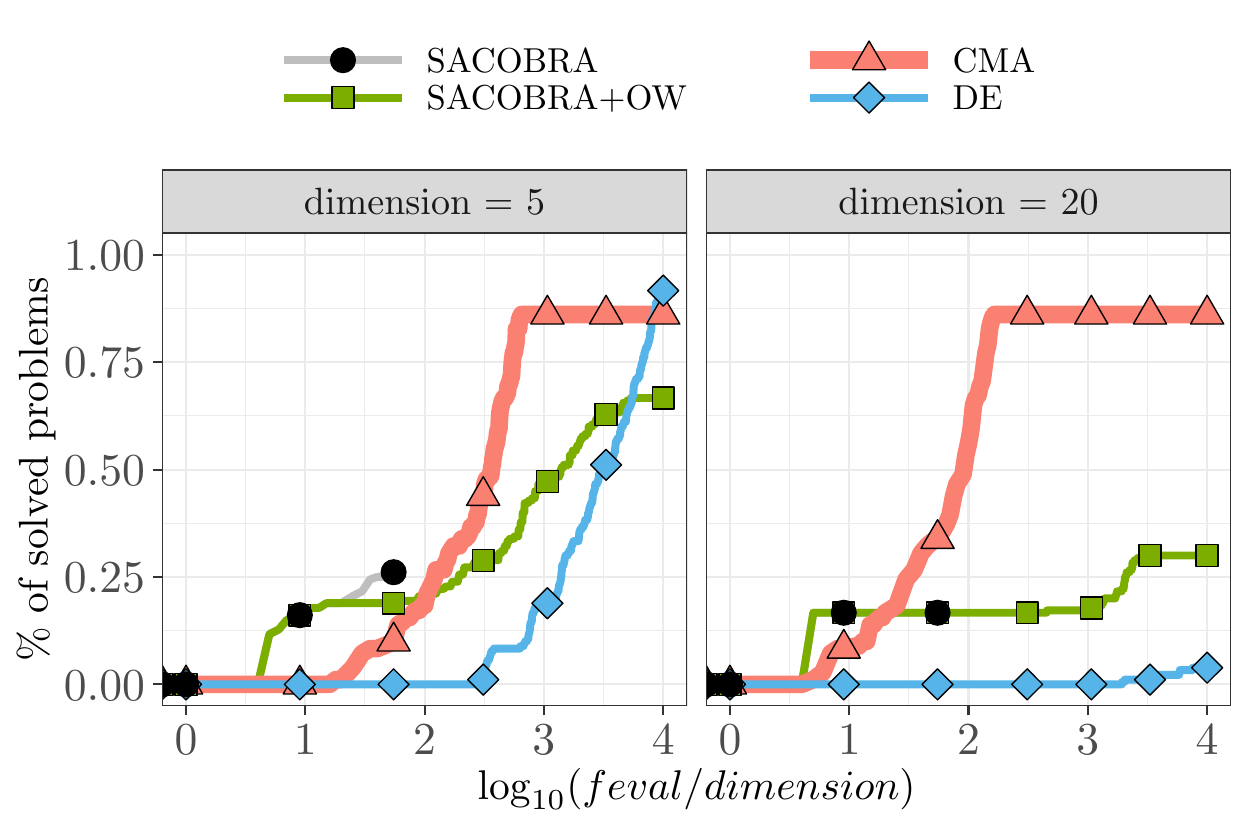}
\caption{Same as Fig.~\ref{fig:DP}, but now for dimension $D=5$ and $D=20$.
%Data profiles for all 12 studied problems in the 5-dimensional and 20-dimensional case. 
The accuracy level is set to $\tau=0.01$.
}
\label{fig:comp520DP-2}
\end{figure}

Fig.~\ref{fig:comp520DP-2} compares the overall performances of the studied algorithms for the 5- and 20-dimensional case. While the results for the case $D=5$ are similar to $D=10$, the higher-dimensional case $D=20$ shows that SACOBRA and SACOBRA+OW as well as DE deteriorate notably. However, CMA-ES stays robust and performs best regardless of the dimensionality. 

%%%%%%%%%%%%%%%%%%%%%%%%%%%%%%%%%%%%%%%%%%%%%%%%
\section{Conclusion}
\label{sec:conc}
Surrogate-assisted optimizers are very fast solvers for linear or non-linear functions with low condition number. But they have severe difficulties when the function to optimize has a high condition number. 
Although we investigated here in detail only RBFs as surrogate models, we have given theoretical arguments that this holds as well for most types of surrogate models, namely for GP models\footnote{ and we have experimental evidence for GP from other runs not shown here}.

We have proposed with SACOBRA+OW a new surrogate-assisted optimization algorithm with online whitening (OW) which aims at transforming online a high-conditioning into a low-conditioning problem. The method OW is applicable to all types of surrogates, not only to RBFs.

The results are encouraging in the sense that SACOBRA+OW finds better solutions than SACOBRA with the same population size. The percentage of solved problems on a subset of the BBOB benchmark is more than doubled when enhancing SACOBRA with OW. 

Although for large budgets ($1000D$ function evaluations and more) SACOBRA+OW outperforms DE, it can no longer be considered as an optimizer for truly expensive problems because of the large number of function evaluations needed for determining the Hessian matrix. While SACOBRA is better for less than $100D$ function evaluations, CMA-ES finds consistently better solutions beyond this point, if we compare by number of function evaluations. But if we have the possibility for parallel computing of the Hessian matrix, then, if we compare by number of iterations,  SACOBRA+OW appears to be the most efficient optimizer among the tested ones. In theory it is always possible to compute a Hessian matrix in parallel but in practice parallelizing this procedure is restricted to the amount of available resources. For example, if  the objective function to optimize is evaluated through a time-expensive simulation run, then $4D+4D^2$ computational cores running in parallel will be required for determining the Hessian matrix in one call. This can be an unrealistic demand when the number of dimensions $D$ is higher.

Another limitation of SACOBRA+OW is that it currently only works well for dimensions $D\leq 20$. 
%\WK{Should we add this remark on $D$?}\SB{Yes, sure}

%{For very large budgets ($1000D$ function evaluations and more), CMAES outperforms SACOBRA+OW. %, while SACOBRA+OW outperforms DE. But SACOBRA+OW outperforms all other investigated algorithms in the expensive setting with a strictly limited budget ($<100D$).}
%\SB{
%\begin{itemize}
	%\item Surrogates may have difficulty to provide reasonable models for functions with high condition number
	%\item SACOBRA+OW finds better solutions in comparison with SACOBRA with the same computational cost but more real function evaluations
	%\item SACOBRA and SACOBRA+OW outperform other algorithms in expensive setting with strictly limited budget
	%\item SACOBRA+OW cannot outperform CMAES when the budget is large enough
%\end{itemize}}

We plan to investigate whether a combination of CMA-ES and surrogate-assisted optimizers could lead to an optimizer which combines 'the best of both worlds'. 
The efficient Hessian estimation of FOCAL~\cite{Shir2011,Shir2014efficient} might be an interesting starting point for this. 

\section{Appendix}
\appendix
\section{Derivation of the Transformation Matrix}\label{sec:append}

Let us assume that the objective function $f(\vec{x})$ is continuous and at least two times differentiable. Its Hessian (matrix of second derivatives)  is $\frac{\partial^2 f(\vec{x})}{\partial \vec{x}^2}=\mathbf{H}$. Here and in the following all partial derivatives are meant to be evaluated at $\vec{x} = \vec{x}_c$, but we suppress this for better readability. $\vec{x}_c$ is the transformation center defined in Eq.~\eqref{eq:transformation0}.
%Let us assume $\vec{x}_c=0$ without loss of generality. 

We show that there is a transformation matrix $\mathbf{M}$ in such a way that the new function $g(\vec{x})=f(\mathbf{M}(\vec{x}-\vec{x}_c))$ becomes spherical, so that its Hessian is 
$\frac{\partial^2 g(\vec{x})}{\partial \vec{x}^2}=\mathbf{I}$. We calculate the derivatives as:

\begin{eqnarray}
\frac{\partial g(\vec{x})}{\partial \vec{x}} &=& \frac{\partial f(\vec{u})}{\partial \vec{x}}\\
                                             &=& \frac{\partial f(\vec{u})}{\partial \vec{u}} \cdot \frac{\partial \vec{u}}{\partial \vec{x}} \\
																						 &=& \frac{\partial f(\vec{u})}{\partial \vec{u}} \cdot \mathbf{M}^T,
\end{eqnarray}

where $\vec{u}=M (\vec{x}-\vec{x}_c)$ and hence $\frac{\partial \vec{u}}{\partial \vec{x}} = \frac{\partial (M (\vec{x}-\vec{x}_c))}{\partial \vec{x}} = \mathbf{M}^T$. 

\begin{eqnarray}
\frac{\partial^2 g(\vec{x})}{\partial \vec{x}^2} &=& \frac{\partial (\frac{\partial f(\vec{u})}{\partial \vec{u}} \cdot \mathbf{M}^T)}{\partial \vec{x}}\\
                                             &=& \frac{\partial (\frac{\partial f(\vec{u})}{\partial \vec{u}} \cdot \mathbf{M}^T)}{\partial \vec{u}} \cdot \frac{\partial \vec{u}}{\partial \vec{x}}\\
																						 &=& \frac{\partial (\frac{\partial f(\vec{u})}{\partial \vec{u}} \cdot \mathbf{M}^T)}{\partial \vec{u}} \cdot \mathbf{M}^T
\end{eqnarray}

We abbreviate $\frac{\partial f(\vec{u})}{\partial \vec{u}}=\vec{P}(\vec{u})$ and can derive 

\begin{eqnarray}
\frac{\partial^2 g(\vec{x})}{\partial \vec{x}^2} &=& \frac{\partial \vec{P} \mathbf{M}^T}{\partial \vec{P}} \cdot \frac{\partial \vec{P}}{\partial \vec{u}}  \cdot \mathbf{M}^T\\
                                                 &=& \mathbf{M} \cdot \frac{\partial^2 f(\vec{u})}{\partial \vec{u}^2} \cdot \mathbf{M}^T\\
																								 &=& \mathbf{M} \cdot \mathbf{H}\cdot \mathbf{M}^T
\end{eqnarray}

We want to ensure that $\frac{\partial^2 g(\vec{x})}{\partial \vec{x}^2}=\mathbf{I}$:\footnote{Strictly speaking, this can only be guaranteed if $g(\vec{x})$ is convex in $\vec{x}_c$. If $g(\vec{x})$ is concave in one or all dimensions, we have a saddle point or local maximum at $\vec{x}_c$. In this case, $\mathbf{I}$ has to be replaced by a diagonal matrix with some elements being $-1$ instead of $1$. But the overall whitening argument remains the same.}

\begin{eqnarray}
\mathbf{I}                          &=& \mathbf{M} \cdot \mathbf{H}\cdot \mathbf{M}^T \label{eq:IHM}\\
\mathbf{M}^{-1}                     &=&  \mathbf{H}\cdot \mathbf{M}^T\\
\mathbf{M}^{-1} (\mathbf{M}^T)^{-1} &=&  \mathbf{H}\\
\mathbf{M}^{T} \mathbf{M}           &=&  \mathbf{H}^{-1}
\end{eqnarray}

A possible solution for the last equation is $\mathbf{M}=\mathbf{H}^{-0.5}$. 

\section{Calculation of Inverse Square Root Matrix}\label{sec:invsquareroot}
We calculate the inverse square root matrix in a numerically stable way with the help of singular value decomposition (SVD)~\cite{Press2007numerical}. The symmetric matrix $\mathbf{H}$  has the SVD representation 

\begin{equation}\label{eq:SVD}
 \mathbf{H} = \mathbf{U}\mathbf{D}\mathbf{V}^T
\end{equation}
with orthogonal matrices $\mathbf{U}$,$\mathbf{V}$ and diagonal matrix $\mathbf{D} = \mbox{diag}(d_i)$ containing only non-negative singular values $d_i$. The inverse square root of $\mathbf{D}$ is 
\begin{equation}\label{eq:invsquareroot}
 \mathbf{D}^{-0.5} = \mbox{diag}(e_i) \quad \mbox{with}\quad e_i = \left\{ \begin{array}{l}
																																		\frac{1}{\sqrt{d_i}} \quad \mbox{if}\quad d_i>10^{-25} \\
																																		0 \qquad \mbox{else}
																																	 \end{array}\right.
\end{equation}

If we define
\begin{equation}\label{eq:M}
 \mathbf{M} = \mathbf{D}^{-0.5} \mathbf{V}^T
\end{equation}
and use the fact that a positive-semidefinite $\mathbf{H}$ has $\mathbf{U} = \mathbf{V}$, then it is easy to show that plugging this $\mathbf{M}$ into Eq.~\eqref{eq:IHM} fulfills the equation.


\begin{thebibliography}{10}

\bibitem{Bagheri2017b}
Samineh Bagheri, Wolfgang Konen, Richard Allmendinger, J\"{u}rgen Branke,
  Kalyanmoy Deb, Jonathan Fieldsend, Domenico Quagliarella, and Karthik
  Sindhya.
\newblock Constraint handling in efficient global optimization.
\newblock In {\em Proc. Genetic and Evolutionary Computation Conference
  GECCO'17}, pages 673--680, New York, 2017. ACM.

\bibitem{Bagheri2017c}
Samineh Bagheri, Wolfgang Konen, and Thomas B\"{a}ck.
\newblock Comparing {Kriging} and radial basis function surrogates.
\newblock In Frank Hoffmann and Eyke H\"{u}llermeier, editors, {\em Proc. 27.
  Workshop Computational Intelligence}, pages 243--259. Universit\"{a}tsverlag
  Karlsruhe, November 2017.

\bibitem{Bagheri2017}
Samineh Bagheri, Wolfgang Konen, Michael Emmerich, and Thomas B\"{a}ck.
\newblock Self-adjusting parameter control for surrogate-assisted constrained
  optimization under limited budgets.
\newblock {\em Applied Soft Computing}, 61:377 -- 393, 2017.

\bibitem{Bagheri2015a}
Samineh Bagheri, Wolfgang Konen, Christophe Foussette, Peter Krause, Thomas
  B{\"a}ck, and Patrick Koch.
\newblock {SACOBRA}: Self-adjusting constrained black-box optimization with
  {RBF}.
\newblock In Frank Hoffmann and Eyke H\"{u}llermeier, editors, {\em Proc. 25.
  Workshop Computational Intelligence}, pages 87--96. Universit\"{a}tsverlag
  Karlsruhe, 2015.

\bibitem{Bajer2015}
Luk\'{a}\v{s} Bajer, Zbyn\v{e}k Pitra, and Martin Hole\v{n}a.
\newblock Benchmarking {Gaussian} processes and random forests surrogate models
  on the {BBOB} noiseless testbed.
\newblock In {\em Proc. Genetic and Evolutionary Computation Conference
  GECCO'15}, pages 1143--1150, New York, 2015. ACM.

\bibitem{Bates1988}
Douglas~M. Bates and Donald~G. Watts.
\newblock {\em Nonlinear regression analysis and its applications}.
\newblock Wiley series in probability and mathematical statistics. Wiley, New
  York [u.a.], 1988.

\bibitem{Bhattacharjee2016}
Kalyan~Shankar Bhattacharjee, Hemant~Kumar Singh, and Tapabrata Ray.
\newblock Multi-objective optimization with multiple spatially distributed
  surrogates.
\newblock {\em Journal of Mechanical Design}, 138(9):091401, 2016.

\bibitem{Fink2009}
Steffen Finck, Nikolaus Hansen, Raymond Ros, and Anne Auger.
\newblock Real-parameter black-box optimization benchmarking 2009: Presentation
  of the noiseless functions.
\newblock Technical Report 2009/20, Research Center PPE, 2009.

\bibitem{Hansen1996}
Nikolaus Hansen and Andreas Ostermeier.
\newblock Adapting arbitrary normal mutation distributions in evolution
  strategies: The covariance matrix adaptation.
\newblock In {\em Proc. of 1996 {IEEE} International Conference on Evolutionary
  Computation, Nayoya University, Japan}, pages 312--317, 1996.

\bibitem{Hansen2011}
Nikolaus Hansen, Raymond Ros, Nikolas Mauny, Marc Schoenauer, and Anne Auger.
\newblock {Impacts of Invariance in Search: When CMA-ES and PSO Face
  Ill-Conditioned and Non-Separable Problems}.
\newblock {\em {Applied Soft Computing}}, 11:5755--5769, 2011.

\bibitem{Jones1998}
Donald~R. Jones, Matthias Schonlau, and William~J. Welch.
\newblock Efficient global optimization of expensive black-box functions.
\newblock {\em J. of Global Optimization}, 13(4):455--492, December 1998.

\bibitem{Kessy2017}
Agnan Kessy, Alex Lewin, and Korbinian Strimmer.
\newblock Optimal whitening and decorrelation.
\newblock {\em The American Statistician}, 2017.
\newblock accepted.

\bibitem{CEC2006}
JJ~Liang, Thomas~Philip Runarsson, Efren Mezura-Montes, Maurice Clerc,
  PN~Suganthan, CA~Coello Coello, and Kalyanmoy Deb.
\newblock Problem definitions and evaluation criteria for the {CEC} 2006
  special session on constrained real-parameter optimization.
\newblock {\em Journal of Applied Mechanics}, 41:8, 2006.

\bibitem{Loshchilov2012a}
Ilya Loshchilov, Marc Schoenauer, and Mich{\`{e}}le Sebag.
\newblock Self-adaptive surrogate-assisted covariance matrix adaptation
  evolution strategy.
\newblock {\em CoRR}, abs/1204.2356, 2012.

\bibitem{WildMore2009}
Jorge~J. Mor\'e and Stefan~M. Wild.
\newblock Benchmarking derivative-free optimization algorithms.
\newblock {\em SIAM J.~Optimization}, 20(1):172--191, 2009.

\bibitem{Posik2012}
Petr Po\v{s}\'{\i}k and V\'{a}clav Klem\v{s}.
\newblock Jade, an adaptive differential evolution algorithm, benchmarked on
  the bbob noiseless testbed.
\newblock In {\em Proceedings of the 14th Annual Conference Companion on
  Genetic and Evolutionary Computation}, GECCO '12, pages 197--204, New York,
  NY, USA, 2012. ACM.

\bibitem{Press2007numerical}
William~H Press.
\newblock {\em Numerical recipes 3rd edition: The art of scientific computing}.
\newblock Cambridge university press, 2007.

\bibitem{Price2005}
Kenneth Price, Rainer Storn, and Jouni~A. Lampinen.
\newblock {\em Differential Evolution: A Practical Approach to Global
  Optimization}.
\newblock Natural Computing Series. Springer, 2005.

\bibitem{Regis2014}
Rommel~G. Regis.
\newblock Constrained optimization by radial basis function interpolation for
  high-dimensional expensive black-box problems with infeasible initial points.
\newblock {\em Engineering Optimization}, 46(2):218--243, 2014.

\bibitem{Regis2015}
Rommel~G. Regis.
\newblock Trust regions in surrogate-assisted evolutionary programming for
  constrained expensive black-box optimization.
\newblock In Rituparna Datta and Kalyanmoy Deb, editors, {\em Evolutionary
  Constrained Optimization}, pages 51--94. Springer, 2015.

\bibitem{Sawyerr2015}
Babatunde~A Sawyerr, Aderemi~O Adewumi, and M~Montaz Ali.
\newblock Benchmarking rcgau on the noiseless bbob testbed.
\newblock {\em The Scientific World Journal}, 2015, 2015.

\bibitem{Sawyerr2013}
Babatunde~A. Sawyerr, Aderemi~O. Adewumi, and Montaz~M. Ali.
\newblock Benchmarking projection-based real coded genetic algorithm on
  bbob-2013 noiseless function testbed.
\newblock In {\em Proceedings of the 15th Annual Conference Companion on
  Genetic and Evolutionary Computation}, GECCO '13 Companion, pages 1193--1200,
  New York, NY, USA, 2013. ACM.

\bibitem{Saxena2015}
Nizin Saxena, Ashish Tripathi, K.~K. Mishra, and A.~K. Misra.
\newblock Dynamic-pso: An improved particle swarm optimizer.
\newblock In {\em 2015 IEEE Congress on Evolutionary Computation (CEC)}, pages
  212--219, May 2015.

\bibitem{Shir2011}
Ofer~M. Shir, Jonathan Roslund, Darrell Whitley, and Herschel Rabitz.
\newblock Evolutionary {Hessian} learning: Forced optimal covariance adaptive
  learning {(FOCAL)}.
\newblock {\em CoRR (arXiv)}, abs/1112.4454, 2011.

\bibitem{Shir2014efficient}
Ofer~M Shir, Jonathan Roslund, Darrell Whitley, and Herschel Rabitz.
\newblock Efficient retrieval of landscape {Hessian}: Forced optimal covariance
  adaptive learning.
\newblock {\em Physical Review E}, 89(6):063306, 2014.

\bibitem{Sutton2007de}
Andrew~M Sutton, Monte Lunacek, and L~Darrell Whitley.
\newblock Differential evolution and non-separability: using selective pressure
  to focus search.
\newblock In {\em Proceedings of the 9th annual conference on Genetic and
  evolutionary computation}, pages 1428--1435. ACM, 2007.

\end{thebibliography}
\end{document}